# Harnessing the Power of Vibration Motors to Develop Miniature Untethered Robotic Fishes

Chongjie Jiang, Yingying Dai, Jinyang Le, Xiaomeng Chen, Yu Xie, Wei Zhou, Fuzhou Niu, *Member, IEEE,* Ying Li, and Tao Luo

*Abstract*—Miniature underwater robots play a crucial role in the exploration and development of marine resources, particularly in confined spaces and high-pressure deep-sea environments. This study presents the design, optimization, and performance of a miniature robotic fish, powered by the oscillation of bio-inspired fins. These fins feature a rigid-flexible hybrid structure and use an eccentric rotating mass (ERM) vibration motor as the excitation source to generate high-frequency unidirectional oscillations that induce acoustic streaming for propulsion. The drive mechanism, powered by miniature ERM vibration motors, eliminates the need for complex mechanical drive systems, enabling complete isolation of the entire drive system from the external environment and facilitating the miniaturization of the robotic fish. A compact, untethered robotic fish, measuring $85 \times 60 \times 45$ mm³, is equipped with three bio-inspired fins located at the pectoral and caudal positions. Experimental results demonstrate that the robotic fish achieves a maximum forward swimming speed of 1.36 body lengths (BL) per second powered by all fins and minimum turning radius of 0.6 BL when powered by a single fin. These results underscore the significance of employing the ERM vibration motor in advancing the development of highly maneuverable, miniature untethered underwater robots for various marine exploration tasks.

## I. Introduction

Underwater robots are widely used for various tasks, such as ocean exploration, ecological monitoring, and aquaculture [1-4]. Traditional large-scale underwater robots typically rely on two primary propulsion methods, which are axial propellers [5-7] and bio-inspired fish-like locomotion. While axial propellers are a common propulsion choice, they have significant limitations, including high noise levels, poor stealth, low energy efficiency, and limited maneuverability. Moreover, they are unsuitable for operation in harsh marine environments, as they can generate acoustic noise that disrupts marine life, damage the seabed, or become entangled in underwater vegetation. In contrast, fish and other aquatic organisms have evolved exceptional locomotion capabilities, demonstrating maneuverability and efficiency far superior to axial propellers. Recent years have seen significant progress in developing bio-inspired robots that mimic aquatic organisms. Currently, large-scale bio-inspired underwater robots primarily utilize three propulsion methods, which are body/caudal fin (BCF) propulsion [8-13], median/paired fin (MPF) propulsion [14-17], and hybrid propulsion combining BCF and MPF propulsion [18-21]. However, the actuation systems of these bio-inspired robots often rely on servomotors, steering engines, and complex mechanical transmission system to generate fin or body oscillations. As a result, these designs tend to be bulky and complex, making it difficult for the robots to navigate through narrow spaces such as coral reef cavities, seabed fissures, valleys, or pits. Therefore, the development of a propulsion method that enables robotic miniaturization is particularly critical.

Piezoelectric actuators [22-24], dielectric elastomer actuators [25-27] and HASEL [28] actuators facilitate the miniaturization of the actuators due to their unique operating mechanisms. However, these actuators typically require high-voltage power supplies, which necessitate large power supply and amplifier modules, thereby limiting the integration of the power, control, and drive systems in a compact size. Additionally, the potential risk of electrical leakage makes these systems a safety concern in underwater applications. Bio-inspired robots utilizing magnetic actuators [29] based on electromagnetic principles can operate with low-voltage power supplies, demonstrating the capability of integration and miniaturization. For instances, Kong et al. [30] reported a miniature untethered robotic fish with a body length of 69 mm, a peak speed of 1.23 body lengths (BL) per second and a minimum turning radius of 0.29 BL. Nagpal et al. [31] designed an underwater robot powered by multiple flapping fins, featuring onboard components such as photodiodes and pressure sensors. This robot has a total length of 100 mm and is capable of three-dimensional motion. Eccentric rotating

This work was supported by the National Natural Science Foundation of China under Grant 52205606 and Grant U21A20136, the Guangdong Basic and Applied Basic Research Foundation under Grant 2021A1515110504, the Natural Science Foundation of Fujian Province under Grant 2022J05011, and the Fundamental Research Funds for the Central Universities under Grant 20720230075. *(Corresponding author: Ying Li and Tao Luo)*

Chongjie Jiang, Yingying Dai, Jinyang Le, Xiaomeng Chen, Yu Xie, and Wei Zhou are with the Pen-Tung Sah Institute of Micro-Nano Science and Technology, Xiamen University, Xiamen 361102, China. (e-mail: 1528552071@qq.com; ing09250205@163.com; 1945774863@qq.com; 617190520@qq.com; xieyu@xmu.edu.cn; weizhou@xmu.edu.cn).

Fuzhou Niu is with School of Mechanical Engineering, Suzhou University of Science and Technology, Suzhou 215009, China. (e-mail: fzniu@usts.edu.cn).

Ying Li is with the School of Mechanical and Electrical Engineering, Shenzhen Polytechnic University, Shenzhen 518055, China. (e-mail: liying@szpu.edu.cn).

Tao Luo is with the Pen-Tung Sah Institute of Micro-Nano Science and Technology, Xiamen University, Xiamen 361102, China, and also with the Shenzhen Research Institute of Xiamen University, Shenzhen 518000, China. (e-mail: luotao@xmu.edu.cn).

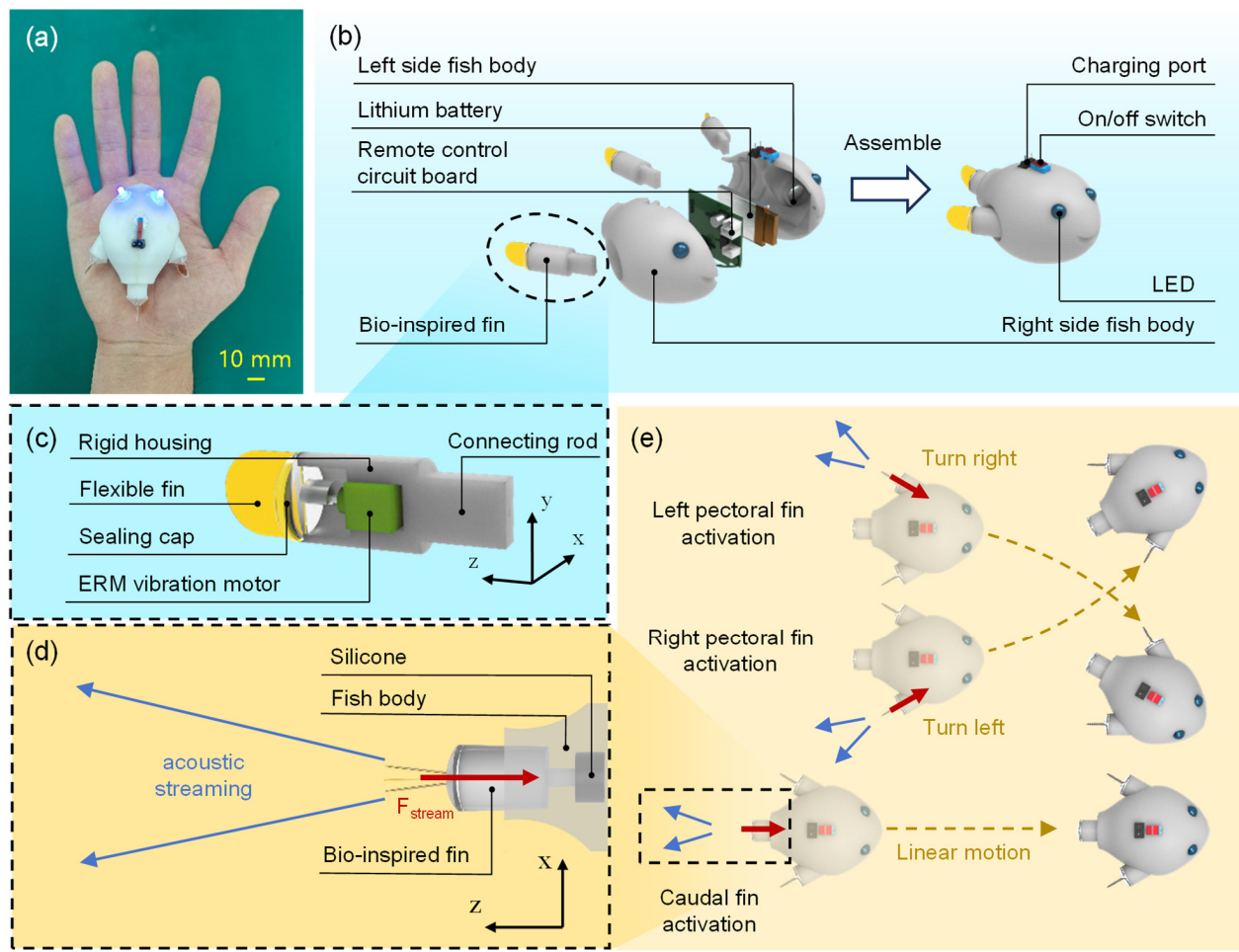

Fig. 1. Design and working principle of the bio-inspired robotic fish. (a) Overall dimensions of the robotic fish. (b) Assembly diagram and components of the robotic fish. (c) Assembly diagram and components of the bio-inspired fin. (d) Propulsion generated by the oscillating bio-inspired fin through acoustic streaming. (e) Multi-modal locomotion of the robotic fish through selective activation of different fin

mass (ERM) vibration motors, also based on electromagnetic principles, provide a simpler but reliable propulsion solution.

These motors require only low-voltage direct current to drive the continuous rotation of an eccentric wheel, generating inertial forces. This mechanism has been successfully applied in the development of an amphibious robot capable of dual locomotion modes of running and swimming [32]. By utilizing rigid-flexible hybrid modules, the flexible fins generate efficient forward thrust by striking the water surface in its third mode, enabling the robot to perform high-speed locomotion on the water surface [33]. However, this work did not demonstrate the underwater locomotion performance of the robot driven by ERM vibration motors. Moreover, the design requires the flexible fin to have oscillation perpendicular to the water surface. However, the x-y direction symmetric, square prism shaped connecting rod of the rigid-flexible hybrid module generates uniform circumferential motion, resulting in a certain degree of efficiency loss.

In this work, we designed a rigid-flexible hybrid bio-inspired fin powered by an ERM vibration motor as the excitation source. In our design, the flexible fin is oriented perpendicular to the rotation plane of the ERM vibration motor, offering several advantages. First, as the eccentric wheel rotates, the flexible fin generates high-frequency, unidirectional vibrations under low-voltage conditions, inducing acoustic streaming that provides thrust perpendicular to the motor's rotation plane, ensuring that the eccentric force does not interfere with the robot's propulsion. Second, we optimized the geometric parameters of the bio-inspired fin to enhance the vibration in the direction perpendicular to the fin's surface, thereby improving underwater propulsion efficiency. Additionally, we developed a remotely controlled bio-inspired robot capable of agile underwater locomotion.

The remainder of this paper is organized as follows: Section II provides an overview of the design of the bio-inspired robot, including its overall structure, bio-inspired fin, and actuation principle. Section III details the optimization of the geometric parameters of the bio-inspired fin through simulations and experiments, examining how these parameters affect propulsion performance. Section IV discusses the robot's locomotion performance, including straight-line motion, turning capabilities, and obstacle avoidance, based on experimental demonstrations. Finally, Section V concludes the study and outlines future works.

## II. DESIGN OF THE ROBOTIC FISH

### A. Structural Design and Locomotion

We designed an untethered miniature bio-inspired robot capable of autonomous underwater locomotion. As shown in Fig. 1(a), the robot's overall dimensions, including flexible fins, are 85 mm × 55 mm × 45 mm, with a total mass of 88 g. Figure 1(b) illustrates the structural components of the robot. The exterior model features an elliptical profile to reduce hydrodynamic drag. Internally, the robot is equipped with a small lithium battery (3.7 V, 55 mAh) that powers all electronic components, along with a remote-control circuit board. The robot's top surface houses a switch, charging interface, and two blue LEDs arranged longitudinally to mimic the eyes of the robotic fish. Three bio-inspired fins are mounted at the pectoral and caudal positions, connected to the exterior model via flexible silicone as a damper to prevent vibration coupling between the fins. All structural components are manufactured using 3D printing, with assembling gaps sealed by waterproof adhesive.

As shown in Fig. 1(c), the bio-inspired fin, consisting of a rigid housing, a flexible fin, a sealing cap and an ERM vibration motor, is the core component responsible for underwater propulsion. The square-shaped body of the vibration motor is glued within the rigid housing, which contains a cylinder-shaped internal cavity allowing the eccentric rotor to rotate freely without interference. As illustrated in Fig. 1(d), the ERM vibration motor generates eccentric force during operation, driving the flexible fin to oscillate at high frequencies and create acoustic streaming to induce thrust force $F_{stream}$ for the propulsion. By controlling the activation and deactivation of different bio-inspired fins, the robot can perform straight-line motion as well as agile left and right turns Fig 1(e).

### B. Design of the Bio-Inspired Fin

A bio-inspired fin in disassembled and assembled states is shown in Fig. 2. The assembly process includes the following steps: (i) the ERM vibration motor (c) is installed into a groove inside the rigid housing (b) and glued with adhesive, ensuring the eccentric rotor can rotate without contacting the housing. (ii) The sealing cap (d) is glued to the rigid housing to fully enclose the ERM vibration motor inside. (iii) One end of the flexible fin (d) is trimmed into a semicircular shape, inserted into a groove on the outer surface of the sealing cap, and fixed with adhesive. The sealing cap is fabricated using PDMS molding, and the flexible fin is made from a 200 μm-thick polyimide film. Note that the connecting rod shown in Fig. 2(a) is used to connect the rigid housing and the body of the robotic fish. The assembled bio-inspired fin (excluding the flexible fin) has an overall length of approximately 28.5 mm, which is small enough to facilitate the miniaturization of the entire robotic fish.

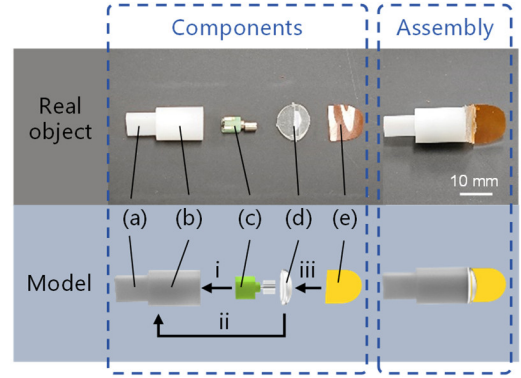

Fig. 2. Rigid-flexible hybrid bio-inspired fin. (a) Connecting rod. (b) Rigid housing. (c) A ERM vibration motor. (d) Sealing cap. (e) Flexible fin.

### C. Actuation Principle

The rotational motion of the eccentric rotor resulted in the oscillation of the flexible fin. As shown in Fig. 3(a), the eccentric rotor completes a full rotation around its axis within a cycle, driving the flexible fin to perform multi-directional harmonic motion. During high-frequency oscillation, the fin's out-plane oscillation significantly disturbs the fluid, generating acoustic streaming.

As shown in Fig. 3(b), the rotating eccentric mass of the ERM vibration motor produces centrifugal force during operation, modeled dynamically as:

$$F(t) = F_0 e^{i\omega t}$$
$$F_0 = md\omega^2$$

where $F(t)$ is the time-dependent centrifugal force, $F_0$ is its amplitude, $m$ is the mass of the eccentric rotor, $d$ is the eccentric distance, and $\omega$ is the angular velocity. The force components along the horizontal $F_x$ and vertical $F_y$ directions are:

$$F_x(t) = F_0 \cos(\omega t)$$
$$F_y(t) = F_0 \sin(\omega t)$$

As shown in Fig. 3(c), the bio-inspired fin can be modeled as a rigid part of the bio-inspired fin connected to a spring (silicone) at one end (the rigid part consists of the connecting rod, the rigid housing, and the ERM vibration motor) and a flexible part of the bio-inspired fin attached to the opposite end of the rigid part (the flexible part consists of the sealing cap and the flexible fin). Under the action of the centrifugal force, the rigid part exhibits a oscillation amplitude $A_{x1}$. Simultaneously, the flexible part, with one end fixed to the rigid part, undergoes deformation, resulting in a deformation amplitude $A_{x2}$ at its free end. The total displacement $A_{x3}$ of the bio-inspired fin can be described as:

$$A_{x3} = A_{x1} + A_{x2}$$

The relationship between acoustic streaming velocity $U$ and oscillation amplitude is given by [34-37]:

$$U \sim \frac{\rho A_{x3}^2 \omega^2}{\mu}$$

where $\rho$ is the fluid density, $\mu$ is the dynamic viscosity. Similarly, the relationship between acoustic streaming thrust $F_{stream}$ and streaming velocity $U$ is:

$$F_{stream} \sim \rho U^2 L_f$$

where $L_f$ is the effective length of the flexible fin. To enhance $F_{stream}$, the amplitudes $A_{1x}$ and $A_{2x}$ must be maximized to achieve the maximum $A_{3x}$.

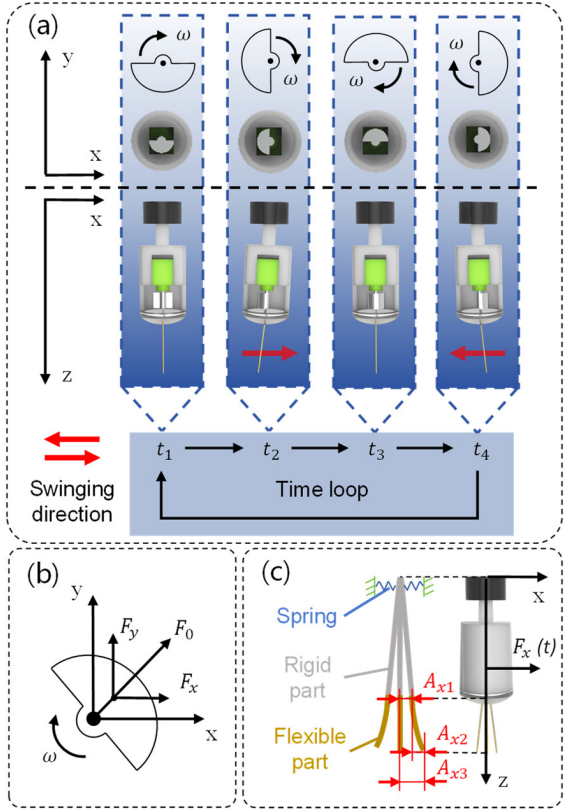

Fig. 3. Driving principle and theoretical model of bio-inspired fin. (a) Actuation principle of the bio-inspired fin. (b) Dynamic model of the rotation of the eccentric mass. (c) Equivalent mechanical model of the bio-inspired fin.

## III. OPTIMIZATION OF THE BIO-INSPIRED FIN

### A. Rigid Part of the Bio-inspired Fin

As shown in Fig. 4(a), the length $L$, height $H$, and width $W$ of the connecting rod jointly influence the vibration modes of the rigid part of the bio-inspired fin, when other dimensions ($L_h$=15.5 mm, $L_s$=5 mm and $D$=11 mm) are fixed. Due to the complexity of analyzing the resonance of the rigid part using analytical methods, we employ the finite element analysis for its modal analysis. As shown in Fig. 4(b), the first mode and second mode shapes of the rigid part correspond to harmonic vibrations along the x-axis and y-axis, respectively. As illustrated in Fig. 4(c), the length $L$ of the connecting rod is correlated with the first natural frequency. When $L$ increases, the first natural frequency of the rigid part decreases. We selected a connecting rod with $L$=10 mm to match the first-mode natural frequency with the rated operating frequency (138 Hz) of the ERM vibration motor. As shown in Fig. 4(d), we conducted finite element analysis for different $H@W$ combinations, and we found that keeping $H \times W$ constant, the first natural frequency of the rigid part remains almost unchanged. When the ratio of $H/W$ increases, the frequency gap between the first and second modes of the rigid part becomes larger, making the rigid part more likely to exhibit harmonic vibration along the x-axis at the first resonance frequency while suppresses its tendency to vibrate harmonically along the y-axis.

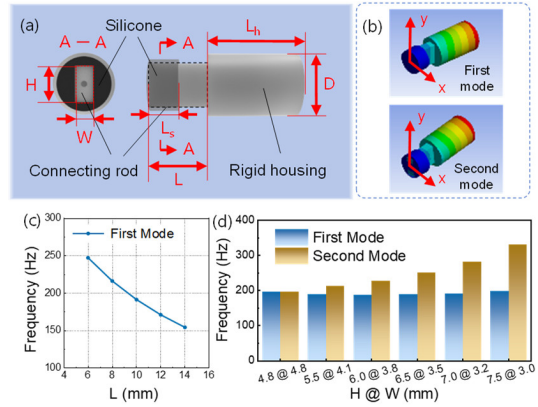

Fig. 4. Modal analysis of the bio-inspired fin. (a) Geometry parameters of the rigid part of the bio-inspired fin. (b) First and second mode shapes of the rigid part of the bio-inspired fin. (c) First resonance frequency of the the rigid part of the bio-inspired fin under different $L$ values. (d) Resonance frequencies of the first and second modes of the rigid part of the bio-inspired fin under different $H/W$ ratios.

As shown in Fig. 5(a), we designed an experiment to validate the effect of the ratio $H/W$ on the vibration of the rigid part of the bio-inspired fin. The connecting rod of the rigid part was connected to a rigid connecting component using elastic silicone adhesive to simulate the real-world scenario of the fish body model. The rigid connecting component was then firmly attached to a fixed tabletop using solid bolt and nut. As shown in Fig. 5(b), the assembled actual experimental setup for studying the effect of the ratio H/W on the vibration of the rigid part of the bio-inspired fin.

In the experiment, as shown in Fig. 5(c), black markers were applied to one side of the rigid housing's exterior surface along the x-directions and y-directions. The initial distances between the two markers, $d_{x0}$ and $d_{y0}$, were measured and recorded via image processing. After activating the ERM vibration motor, a camera was used to capture the vibration process. Image processing was then employed to determine the distances $d_x$ and $d_y$ caused by the vibrations in the x-directions and y-directions, respectively. In this way, the maximum amplitudes in the horizontal $A_{1x}$ and vertical $A_{1y}$ directions were calculated as:

$$A_{1x} = d_x - d_{x0}$$
$$A_{1y} = d_y - d_{y0}$$

In the optimization design, as shown in Fig. 5(d), when the driving voltage of the ERM vibration motor increases, its vibration frequency also increases. According to Fig. 5(e) and Fig. 5(f), as the voltage increases, the amplitude of the rigid part in both the x- and y-directions increases. When the height $H$ and width $W$ are equal, the amplitudes $A_{1x}$ and $A_{1y}$ are approximately equal at the same voltage. Notably, as shown in Fig. 5(g), the larger the ratio of $H/W$, the greater the ratio of $A_{1x}/A_{1y}$, indicating a more pronounced movement trend of the rigid part in the x-direction. Therefore, adopting a higher $H/W$ ratio helps achieve the desired oscillation direction. Considering the overall size constraints of the robotic fish, the connecting rod of the rigid part was designed with dimensions of height $H$=7.5 mm and width $W$=3 mm.

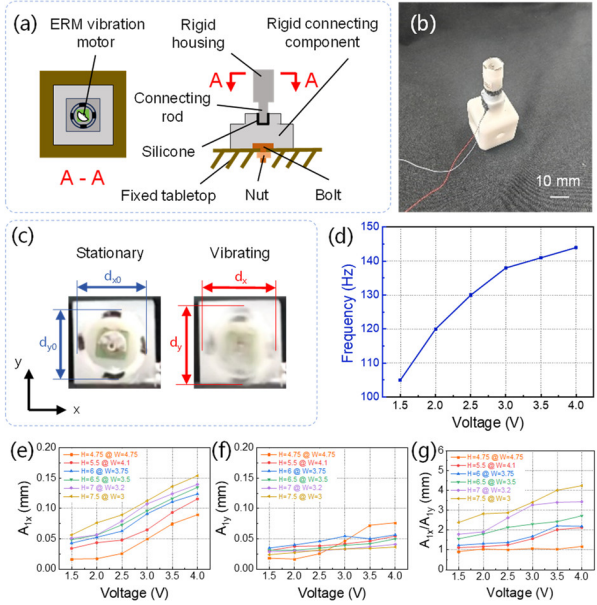

Fig. 5. Experiment to validate the vibration mode of the rigid part. (a) Schematic of the experimental setup for characterizing the oscillation of bio-inspired fins. (b) Photograph of the experimental apparatus. (c) Imaging of the stationary (left) and oscillating (right) rigid part of the bio-inspired fin. (d) Vibration frequency of the ERM vibration motor under different voltages. (e) Amplitudes in the x-direction under various $H/W$ ratios. (f) Amplitudes in the y-direction under various H/W ratios. (g) Ratios of $A_{1x}/A_{1y}$ under various H/W ratios.

## B. Flexible Fin

The robotic fish employs high-frequency, unidirectional oscillations of bio-inspired fins powered by ERM vibration motors to achieve underwater locomotion. As the bio-inspired fin oscillates underwater, its motion generates fluid flow that propels the robotic fish. As shown in Fig. 6(a), we designed flexible fins of varying lengths and analyzed their first mode of oscillation. The oscillation of the entire fin combines the free-end motion of its rigid part with the deformation of the flexible fin. Longer flexible fins reduce the natural frequency of the first-mode oscillation of the entire bio-inspired fin. In this study, we aim to achieve a bio-inspired fin with a first-mode natural frequency that closely matches the rated frequency of the vibration motor.

A thrust testing platform was constructed to evaluate the propulsion performance of the bio-inspired fin. As shown in Fig. 6(b), the platform consists of a rigid connecting component, a force sensor, and a floating boat. The bio-inspired fin is affixed to one end of the connector using flexible silicone, while the other end of the connector is bolted to the floating boat. The floating boat is tethered to a tension sensor via thin strings, and the bio-inspired fin is connected to the rigid connecting component via elastic silicone. When the bio-inspired fin is actuated, high-frequency oscillations generate acoustic streaming, propelling the floating boat in the opposite direction and applying a tensile force on the force sensor. The measured force $F_{tension}$ was used to represent the fin's thrust $F_{stream}$. Figure 6(c) shows the $F_{tension}$ versus driving voltage $V$ for different fin lengths. In the voltage range of 3 to 4 V (corresponding to a vibration frequency of 138–144 Hz), fins with lengths $L_f$ = 9 mm and 12 mm produced greater $F_{stream}$ than fins with $L_f$ = 6 mm, 15 mm, or 18 mm. The possible reasons are: (i) For fins with lengths $L_f$ = 6mm, 9mm and 12mm, the frequency of the ERM vibration motor matches the natural frequency of the bio-inspired fin, allowing resonance to occur; (ii) Oscillation amplitude $A_{x3}$ of the flexible fin is positively correlated with its length $L_f$, thus the bio-inspired fin with $L_f$ =6 mm exhibits a smaller $A_{x3}$, resulting in a lower thrust compared to fins with $L_f$ = 9 mm and 12 mm; (iii)The frequency of the ERM vibration motor exceeds the natural frequencies of the bio-inspired fins with $L_f$ =15 mm and 18 mm, preventing them from achieving resonance.

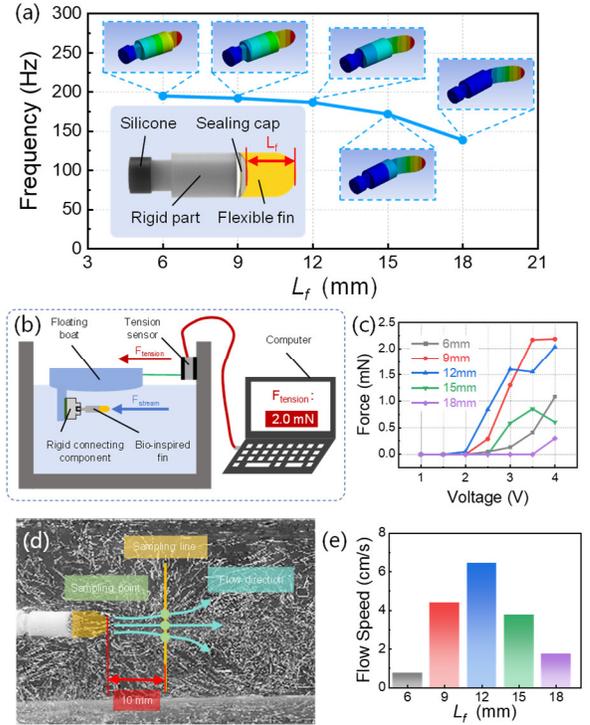

Fig. 6. The driving performance of the bio-inspired fin. (a) Oscillation modes and natural frequencies of the entire bio-inspired fin with different lengths $L_f$ of flexible fins. (b) Schematic of the thrust testing platform for the bio-inspired fin. (c) Thrust of the oscillating bio-inspired fins with different lengths $L_f$ of flexible fins under different voltages. (d) Visualization of the acoustic streaming and its velocity sampling. (e) Acoustic streaming velocity of bio-inspired fins with different lengths $L_f$ at a voltage of 3V.

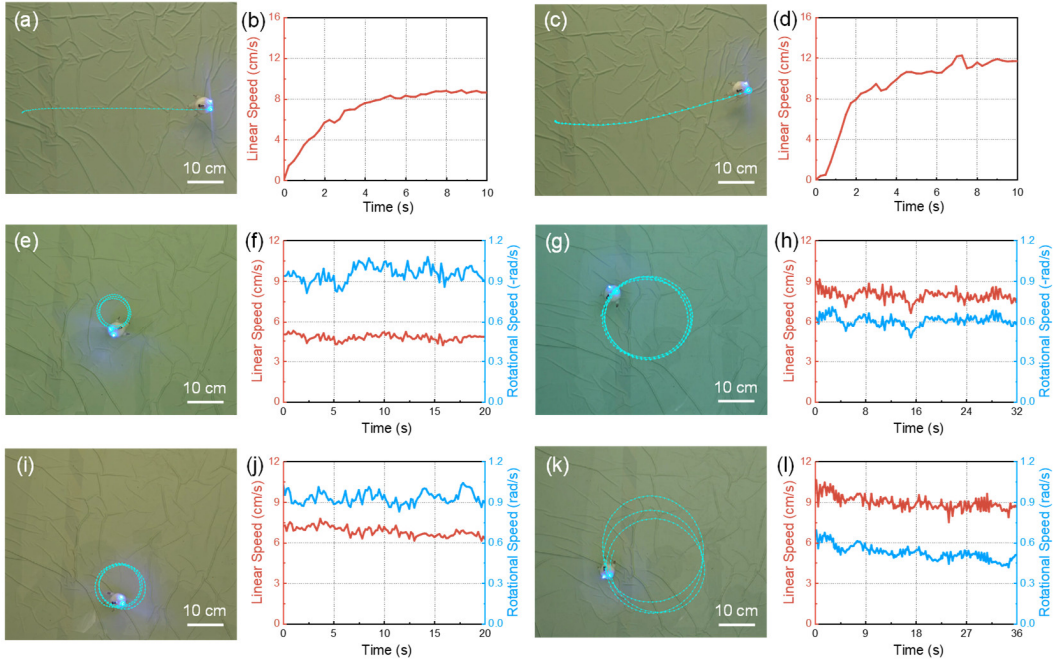

Fig. 7. Locomotion performance testing of the robotic fish. (a) Trajectory of the straight swimming driven by a single caudal fin. (b) Linear velocity of the straight swimming driven by a single caudal fin. (c) Trajectory of the straight swimming driven by the simultaneous activation of all bio-inspired fins. (d) Linear velocity of straight swimming driven by all bio-inspired fins simultaneously. (e) Trajectory of the right turning driven by a single left pectoral fin. (f) Linear velocity and angular velocity of the right turning driven by a single left pectoral fin. (g) Trajectory of the right turning driven by the simultaneous activation of the caudal fin and left pectoral fin. (h) Linear velocity and angular velocity of the right turning driven by the simultaneous activation of the caudal fin and left pectoral fin. (i) Trajectory of the left turning driven by the right pectoral fin. (j) Linear velocity and angular velocity of the right turning driven by the right pectoral fin. (k) Trajectory of the left turning driven by the simultaneous activation of the caudal fin and the right pectoral fin. (l) Linear velocity and angular velocity of the right turning driven by the simultaneous activation of the caudal fin and the right pectoral fin.

## IV. LOCOMOTION OF THE ROBOTIC FISH

Based on the optimized bio-inspired fin equipped with the flexible fin of 12 mm, a minimized fish body that is sufficiently to accommodate the necessary power supply and control components is designed. The axes of rotation for the ERM vibration motor of the two pectoral fins and the caudal fin are positioned on the same plane, with each pectoral fin forming an angle of 30° with the caudal fin. This section demonstrates the underwater locomotion performance of the miniature robotic fish, including locomotion driven by a single bio-inspired fin and multiple fins.

### A. Actuation with A Single Bio-Inspired Fin

Under the rated operating voltage of 3.0 V, swimming tests were conducted in a water tank, and the motion of the robotic fish was recorded with a camera for image processing to determine its velocity characteristics. As shown in Fig. 7(a), the motion trajectory demonstrates that the robotic fish achieves stable forward motion driven by the caudal fin alone. As shown in Fig. 7(b), the robotic fish achieves a steady swimming speed of 8.53 cm/s (approximately 1.0 BL/s) after 5 seconds of acceleration, which reflects the overall performance of the caudal fin.

Activating only the left pectoral fin causes the robot to turn right. Figure 7(e) illustrates the trajectory of a right turn with a turning radius of 6 cm (approximately 0.7 BL). As shown in Fig. 7(f), the robot reaches a stable right-turn speed around 5 cm/s (approximately 0.6 BL/s) with an angular velocity of 1 rad/s. Conversely, activating only the right pectoral fin causes the robot to turn left. Figure 7(i) shows the left-turn trajectory with a turning radius of 7 cm (approximately 0.82 BL). As shown in Fig. 7(j), the robot reaches a stable left-turn speed of 7 cm/s (approximately 0.82 BL/s) with an angular velocity of 1 rad/s.

Slight performance differences between the left and right pectoral fins may result from manufacturing non-uniformity during the manual assembly of the robotic fish and minor deviations of the robot's center of mass.

### B. Actuation with Multiple Fins

As shown in Fig. 7(c), the robot demonstrates enhanced linear motion by activating the caudal fin and both pectoral fins simultaneously. The trajectory indicates an initial straight-line motion followed by a slight deviation due to the slight performance difference between two pectoral fins. As shown in Fig. 7(d), the robotic fish reaches a steady forward swimming speed of 11.6 cm/s (approximately 1.36 BL/s) within 6 seconds.

Activating both the caudal fin and the left pectoral fin causes the robot to turn right. Figure 7(e) shows the trajectory, with a turning radius of 13 cm (approximately 1.52 BL). As shown in Fig. 7(f), the robotic fish achieves a steady right-turn speed of 8.5 cm/s (approximately 1.0 BL/s) with an angular velocity of 0.6 rad/s. Conversely, activating both the caudal fin and the right pectoral fin causes the robot to turn left. Figure 7(i) illustrates the left-turn trajectory, with a turning radius of 15 cm (approximately 1.76 BL). As shown in Fig. 7(j), the robotic fish achieves a stable left-turn speed of 9 cm/s (approximately 1.06 BL/s) with an angular velocity of 0.5 rad/s.

By coordinating the actuation of three bio-inspired fins at different positions, the robotic fish can be controlled to move with different locomotion combinations, which empowers the potential application of the robotic fish in confined underwater environments with high-density obstacles.

### C. Controlled locomotion through coordinated activation of bio-inspired fins.

Figure 8 presents snapshots of the robot's locomotion in different environmental scenarios. As shown in Fig. 8(a), by precisely controlling the activation and deactivation of three bio-inspired fins, the robotic fish is able to navigates swiftly between floating balls in a water tank. Figure 8(b) showcases the robot's ability to perform remote-controlled swimming in a practical outdoor environment, navigating around a water level measurement post in a reservoir. All these results highlight the potential application of the robotic fish in confined spaces with obstacles for underwater exploration, environmental monitoring, and beyond.

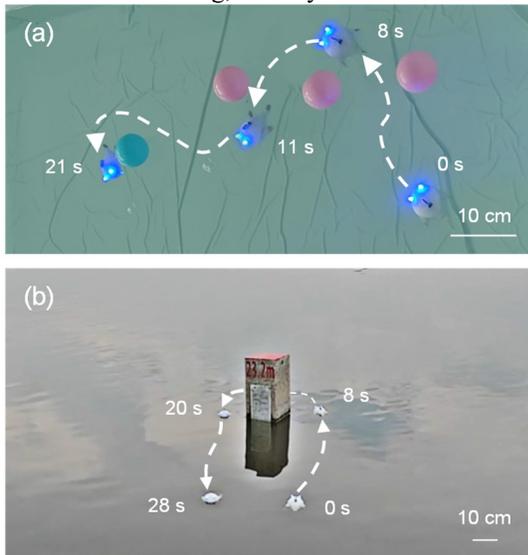

Fig. 8. Controlled locomotion performance of the robotic fish. (a) The robotic fish navigates through narrow channels between multiple floating balls. (b) The robotic fish swims around a water level measurement

## V. CONCLUSIONS AND FUTRUE WORK

We have developed a miniature robotic fish, equipped with three rigid-flexible coupled bio-inspired fins located at the pectoral positions on both sides and the caudal position at the rear. Those bio-inspired fins use ERM vibration motors, which are installed inside rigid housings and fully sealed with flexible fins, as excitation sources. Compared to traditional large-scale underwater robots with rotary shaft actuators, these vibration-driven bio-inspired fins eliminate the need for exposing shafts to the water environment, achieving complete isolation of the excitation source from the external environment, which is intrinsically watertight. In addition, the ERM vibration motor actuated bio-inspired fin does not need complex mechanical drive system, which significantly reducing mechanical complexity and facilities the miniaturization of the entire robotic fish. Experimental results show that the robotic fish has a maximum translational speed of 11.56 cm/s (approximately 1.36 BL/s) and a minimum turning radius of 6 cm (0.7 BL). These results indicate that the miniature robotic fish can achieve precise and versatile locomotion control through coordinated activation of those bio-inspired fins, which empowers its potential for applications such as underwater exploration and environmental monitoring.

Future work will explore the possibility of manufacturing the fish body using soft materials, such as PDMS and Eco-flex silicone rubber, to further reduce vibration coupling between the different bio-inspired fins. Additionally, to overcome the limitations of electromagnetic signal transmission in underwater environments, acoustic communication methods will be investigated to enable reliable remote control of the robot in deep-water conditions.